\documentclass[runningheads]{llncs}
\usepackage[T1]{fontenc}
\usepackage[small]{caption}
\usepackage{graphicx}
\usepackage{amsmath}
\usepackage{algorithm}
\usepackage{algorithmic}
\usepackage[switch]{lineno}
\usepackage{subfigure}
\usepackage{diagbox}
\usepackage{booktabs}
\usepackage{xcolor}
\usepackage{adjustbox}
\usepackage{hyperref}
\begin{document}
%

% \title{Contribution Title\thanks{Supported by organization x.}}
\title{Contrastive Conditional Alignment based on Label Shift Calibration for Imbalanced Domain Adaptation}
\titlerunning{CCA-LSC for Imbalanced Domain Adaptation}
% If the paper title is too long for the running head, you can set
% an abbreviated paper title here
%

% \author{First Author\inst{1}\orcidID{0000-1111-2222-3333} \and
% Second Author\inst{2,3}\orcidID{1111-2222-3333-4444} \and
% Third Author\inst{3}\orcidID{2222--3333-4444-5555}}
% %
% \authorrunning{F. Author et al.}
% % First names are abbreviated in the running head.
% % If there are more than two authors, 'et al.' is used.
% %
% \institute{Princeton University, Princeton NJ 08544, USA \and
% Springer Heidelberg, Tiergartenstr. 17, 69121 Heidelberg, Germany
% \email{lncs@springer.com}\\
% \url{http://www.springer.com/gp/computer-science/lncs} \and
% ABC Institute, Rupert-Karls-University Heidelberg, Heidelberg, Germany\\
% \email{\{abc,lncs\}@uni-heidelberg.de}}
%

\author{Xiaona Sun \and Zhenyu Wu \thanks{Corresponding author. Email: shower0512@bupt.edu.cn} \and Zhiqiang Zhan \and Yang Ji}
\authorrunning{Xiaona Sun \and Zhenyu Wu \and Zhiqiang Zhan \and Yang Ji.}
\institute{School of Communication and Information Engineering
Beijing Univ. of Posts and Telecommunications\\
\email{\{subxiaona,shower0512,zqzhan,jiyang\}@bupt.edu.cn}}

% \author[1]{Xiaona Sun}
% \ead{subxiaona@bupt.edu.cn}
% \author[1]{Zhenyu Wu}
% \cormark[1]
% % [style=chinese]
% \ead{shower0512@bupt.edu.cn}
% % Third author
% \author[1]{Yang Ji}
% \ead{jiyang@bupt.edu.cn}
% \author[1]{Zhiqiang Zhan}

\maketitle              % typeset the header of the contribution
\begin{abstract}
Many existing unsupervised domain adaptation (UDA) methods primarily focus on covariate shift, limiting their effectiveness in imbalanced domain adaptation (IDA) where both covariate shift and label shift coexist. Recent IDA methods have achieved promising results based on self-training using target pseudo labels. However, under the IDA scenarios, the classifier learned in the source domain will exhibit different decision bias from the target domain. It will potentially make target pseudo labels unreliable, and will further lead to error accumulation with incorrect class alignment.
Thus, we propose contrastive conditional alignment based on label shift calibration (CCA-LSC) for IDA, to address both covariate shift and label shift. Initially, our contrastive conditional alignment resolve covariate shift to learn representations with domain invariance and class discriminability, which include domain adversarial learning, sample-weighted moving average centroid alignment and discriminative feature alignment. Subsequently, we estimate the probability distribution of the target domain, and calibrate target sample classification predictions based on label shift metrics to encourage labeling pseudo-labels more consistently with the distribution of real target data. Extensive experiments are conducted and demonstrate that our method outperforms existing UDA and IDA methods on benchmarks with both label shift and covariate shift. Our code is available at \textcolor{blue}{\url{https://github.com/ysxcj-hub/CCA-LSC}}.

% \keywords{First keyword  \and Second keyword \and Another keyword.}
\end{abstract}

\section{Introduction}

Unsupervised Domain Adaptation (UDA) ~\cite{ben2010theory,long2016unsupervised,sun2016deep,long2015learning} aims to transfer knowledge from labeled source domain to unlabeled target domain. A common scenario in UDA is covariate shift, where the conditional distributions of the labels given the features are the same across domains, i.e., $P_S(y|x)=P_T(y|x)$, but the marginal distributions of the features are different, i.e., $P_S(x) \neq P_T(x)$. Many UDA methods have been proposed to deal with covariate shift, such as distribution matching-based methods~\cite{sun2016deep,long2015learning,ganin2016domain}, which aim to align the feature distributions of the source and target domains by minimizing some distance measure. However, when there exists label distribution shift, i.e. $P_S(y|x) \neq P_T(y|x)$, distribution matching-based methods may suffer from negative transfer.
In real-world scenarios, domain adaptation often faces the challenge of both data distribution shift (covariate shift) and label distribution shift (label shift). Moreover, real-world data is usually imbalanced, where some classes are more frequent than others. For example, in the domainnet ~\cite{peng2019moment} dataset, the head classes that are abundant in the source domain may be scarce in the target domain. This scenario is referred to as imbalanced domain adaptation (IDA). To enable domain adaptation to cope with such realistic situations, effective IDA algorithms are essential.

% \begin{figure}
%     \centering
%     \includegraphics[width=0.55\textwidth]{figure/introduction_1.jpg}
%     \caption{ {\bf Left Top:} In cases of significant label shift, class-balanced resampling on source domain may still mislabel target samples due to unknown target label distribution. This can cause error accumulation and incorrect class alignment in IDA methods using self-training with pseudo-labels. {\bf Left Bottom:}Our method rectifies the classification boundary to predict the target samples based on the label shift metric $M_{ls}$, which effectively reduces the error rates of estimating target pseudo labels. And we use calibrated pseudo-labels in contrastive conditional alignment to learn feature representations that are both domain-invariant and class-discriminative.}
%     \label{introduction}
% \end{figure}
% \begin{figure}
%     \centering
%     \includegraphics[width=0.6\textwidth]{figure/dataset_1.jpg}
%     \caption{Label distributions on DomainNet and OfficeHome}
%     \label{dataset}
% \end{figure}

\begin{figure*}[ht]
\centering
\subfigure[]{
\includegraphics[width=0.4\textwidth]{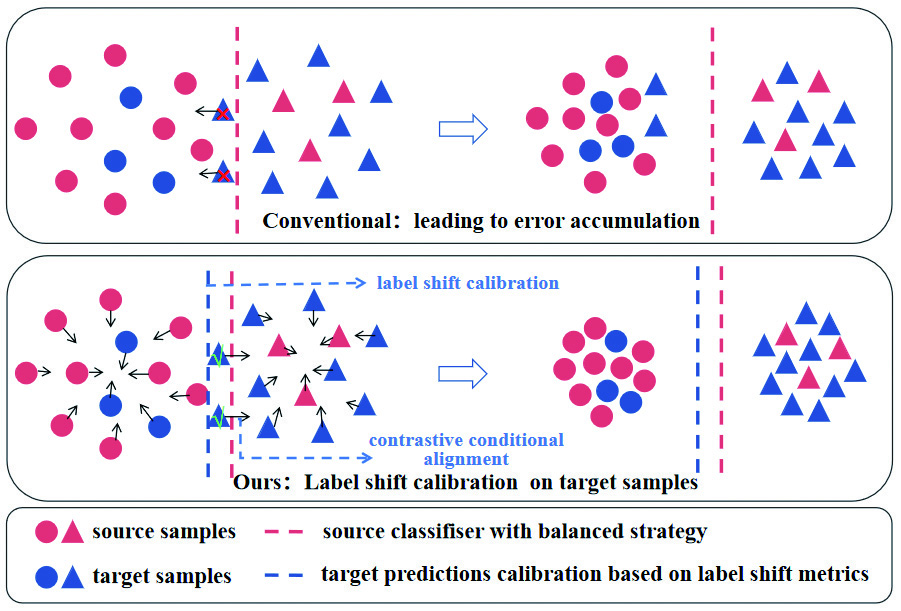}
\label{introduction}
}
\subfigure[]{
\includegraphics[width=0.5\textwidth]{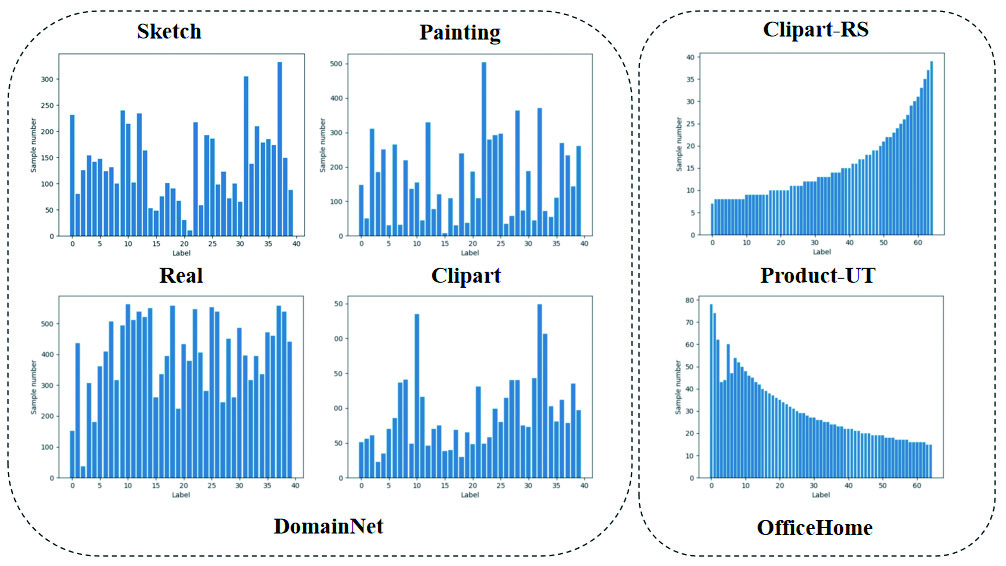}
\label{dataset}
}
\caption{{\bf (a)Top:} In cases of substantial label shift, classifier learned from source domain may  mislabel target samples due to the unknown target label distribution. This can result in error accumulation and misalignment in IDA methods that use self-training with pseudo-labels. {\bf (a)Bottom:} Our approach rectifies the classification boundary to predict target samples based on the label shift metric $M_{ls}$, effectively reducing the error rates in estimating target pseudo-labels. We employ calibrated pseudo-labels in CCA to learn feature representations that are both domain-invariant and class-discriminative. {\bf (b):} Label distributions on DomainNet and OfficeHome}
% \caption{ {\bf (a)Top:} In cases of significant label shift, class-balanced resampling on source domain may still mislabel target samples due to unknown target label distribution. This can cause error accumulation and incorrect class alignment in IDA methods using self-training with pseudo-labels. {\bf (a)Bottom:} Our method rectifies the classification boundary to predict the target samples based on the label shift metric $M_{ls}$, which effectively reduces the error rates of estimating target pseudo labels. And we use calibrated pseudo-labels in contrastive conditional alignment to learn feature representations that are both domain-invariant and class-discriminative. {\bf (b):} Label distributions on DomainNet and OfficeHome}
\label{fig:mainfig2}
\end{figure*}

Recent studies attempt to address the IDA problem through self-training with target pseudo-labels. However, these methods prove unstable as the classifier’s output tends to align more closely with the source than the target label distribution under label shift. This discrepancy results in noisier pseudo-labels for target samples. The issue is particularly pronounced for classes with a large label shift, leading to error accumulation, as depicted in Figure \ref{introduction}, top.

To tackle this issue, we introduce a novel method termed contrastive conditional alignment based on label shift calibration (CCA-LSC). This method adjusts the classification of target samples in accordance with the degree of label shift. First, we propose to align the conditional distributions of two domains inspired by contrastive learning by using domain adversarial learning, sample-weighted moving average centroid alignment, and discriminative feature alignment. We then estimate the label distribution of the target domain ($\widehat P_T$) after a simple pre-training. Second, we utilize $\widehat P_T$ and the label distribution of the source domain $P_S$ to calculate the degree of label shift. Finally, we adjust the classification prediction of target samples according to the degree of label shift during the training process. Our experiments reveal that the pseudo-labels procured by CCA-LSC consistently outperform the pseudo-labels obtained directly from the classifier’s output. This observation, confirmed across all tasks on the OfficeHome and DomainNet datasets, suggests that this strategy effectively enhances the reliability of pseudo-labels, thereby promoting a more accurate alignment across the two domains. See Figure \ref{introduction}, bottom. 

The contributions of this article are as follows:
\begin{itemize}
    % \item We introduce a novel metric to quantify label shift, and for the first time, leverage this metric to rectify the classification predictions of target samples. This approach effectively mitigates the issue of unreliable pseudo-labels induced by label shift.
    \item Contrastive conditional alignment (CCA) leverages the principles of contrastive learning for extracting domain-invariant and class-discriminative features to resist covariate shift. And it weights samples to reduce misalignment from unreliable target pseudo labels.
    \item Label shift calibration (LSC) introduce a novel metric to quantify label shift and leverage this metric to rectify the classification predictions of target samples, which reduce target false pseudo-rate and resist label shift. CCA and LSC jointly resolve the IDA problem.
    % \item  CCA-LSC introduces contrastive conditional alignment to address covariate shift and proposes label shift calibration to tackle label shift.  The latter’s calibrated pseudo-labels are utilized by the former for training in the target domain, jointly resolving the IDA problem.
    \item Experiments were conducted on the OfficeHome and DomainNet datasets, which have both label shift and covariate shift, and it was shown that CCA-LSC achieved state-of-the-art performance.

\end{itemize}

\section{Related Work}
\subsubsection{Unsupervised Domain Adaptation With Covariate Shift}
Covariate shift in UDA is primarily addressed by three kind of methods: statistic divergence alignment, adversarial training, and self-training. Statistic divergence alignment learns invariant features by minimizing domain discrepancy, with the divergence measure selection being key. Measures such as maximum mean discrepancy (MMD) ~\cite{long2016unsupervised,long2015learning,long2017deep,ge2023unsupervised}, correlation alignment ~\cite{sun2016deep}, wasserstein distance ~\cite{balaji2019normalized,lee2018minimax,shen2018wasserstein}, marginal discrepancy measures ~\cite{zhang2019bridging}, and other distance-based methods ~\cite{li2019locality,li2021faster} are commonly employed. Adversarial training, taking inspiration from generative adversarial networks (GANs) ~\cite{goodfellow2014generative}, aims to extract domain invariant features ~\cite{ganin2016domain,long2018conditional,saito2018maximum,rangwani2022closer}  through an adversarial process. These UDA methods align the marginal distribution during training, assuming invariant label distributions. However, label shifts could lead to bad performance or even negative transfer. Self-training ~\cite{mei2020instance,wei2020theoretical,zou2019confidence} employs pseudo-labels generated from the target domain for training on target domain data. However, these pseudo-labels may suffer from miscalibrated probabilities ~\cite{guo2017calibration}, potentially leading to the errors accumulation.

\subsubsection{Unsupervised Domain Adaptation With Label Shift}
These techniques strive to tackle the challenge of varying label distributions across domains. Predominant strategies include class-weighting methods ~\cite{lipton2018detecting,azizzadenesheli2019regularized,azizzadenesheli2021importance} and those that address cross-domain label shift by predicting and estimating the distribution of the target label ~\cite{lipton2018detecting,alexandari2020maximum}. However, these methods presume the feature distribution is invariant across domains, only concentrating on label shift. Additional methods have investigated DA scenarios where the label spaces across domains do not entirely overlap, such as open set domain adaptation ~\cite{panareda2017open,yang2020heterogeneous} and partial domain adaptation ~\cite{cao2018partial,zhang2018importance,zhang2023digital}. These methods pertain to specific label shift problems, which are not the focus of this paper.
\subsubsection{Imbalanced Domain Adaptation}
IDA is designed to tackle the coexistence of covariate shift and label shift. Typical methods include conditional distribution alignment based on pseudo-labels ~\cite{tan2020class,tachet2020domain}, class-weighting strategies ~\cite{wang2017balanced,yan2017mind}, implicit alignment methods based on sampling ~\cite{jiang2020implicit}, asymmetric relaxed distribution alignment ~\cite{wu2019domain}, and cluster-level discrepancy minimization ~\cite{yang2021advancing}. These methods typically utilize pseudo-labels for self-training. However, under strong label shift, pseudo-labels are often unreliable, leading to error accumulation and erroneous class alignment. To address this, SENTRY ~\cite{prabhu2021sentry} proposed that minimizes the entropy of reliable instances and maximizes the entropy of unreliable instances. ISFDA ~\cite{li2021imbalanced} proposed a method using secondary label correction. However, as label shift varies for different classes, unreliable instances are class-biased. These methods overlook the varying label shift across classes and do not essentially address the label shift issue. In this work, we introduce CCA-LSC. It adjusts the classification prediction of target samples based on each class’s label shift degree, $M_{ls}$, enhancing the precision of pseudo-labels.
 % PAT ~\cite{shi2022pairwise} suggests generating adversarial samples from interpolated, aligned source-target pairs. However, unreliable pseudo-labels and misaligned features may cause samples to stray from the decision boundary. 

\section{Method}
\subsection{Problem Setup}
In this work, we investigated C-way image classification. 
In imbalanced domain adaptation (IDA), we are given a source domain $\mathcal{S}=\{\mathcal{X}_i^s,\mathcal{Y}_i^s\} $ with $N_s$ labeled samples $\{(x_i^s,y_i^s)_{i=1}^{N_s}\}$ and a target domain $\mathcal{T}=\{\mathcal{X}_i^t\}$ with $N_t$ unlabeled samples $\{(x_i^t)_{i=1}^{N_t}\}$, where the input $x$ are images and label $y \in \{1,2, \dots,C\}$ are categorical variables. For the joint case of label shift and covariate shift, we adopt the same assumption in ~\cite{tan2020class}, i.e., $p(y|x)=q(y|x)$, $p(x)\neq q(x)$, $p(y)\neq q(y)$ and $p(x|y)\neq(x|y)$. Our goal is to learning a CNN mapping function $f_t$: $\mathcal{X}_t \to \mathcal{Y}_t$.

% In Section 3.2, we initially present contrastive conditional alignment (CCA), which include domain adversarial learning, sample-weighted moving average centroid alignment, and discriminative feature alignment, where the latter two are inspired by contrastive learning. These strategies tackle covariate shift and provide a estimation for target domain’s label distribution via rudimentary pre-training. Section 3.3 delves into addressing the label shift problem. We discuss the computation of label shift degree metrics $M_{ls}$ and introduce the label shift calibration (LSC) on target samples based on $M_{ls}$. Finally, in Section 3.4, we summarize overall optimization and theoretically analyze the effectiveness of our method.

\subsection{Contrastive Conditional Alignment (CCA)}
\subsubsection{Domain Adversarial Learning}
In domain adversarial learning, an auxiliary domain classifier $\mathbf{D}$ is employed to determine whether the features extracted by $\mathbf{G}$ are derived from the source or target domain. Simultaneously, $\mathbf{G}$ is trained to deceive $\mathbf{D}$. When this adversarial game reaches a state of equilibrium, the features produced by G demonstrate domain invariance. Formally,
\begin{align}\label{eq:(1)}
    L_{DC}(x_s, x_t) = E_{x \sim D_S} [\log(1 - D \circ G(x))]  %\nonumber\\
      + E_{x \sim D_T} [\log(D \circ G(x))].
\end{align}
\subsubsection{Sample-weighted Moving Average Centroid Alignment}
However, domain-invariance does not mean cross domain class-invariance. In ~\cite{xie2018learning}, they propose to use moving average centroid alignment strategy. This strategy explicitly constrains the distance between centroids with identical class but different domains, ensuring close mapping of same-class features. The transfer objective is:
\begin{equation}\label{eq:(2)}
    L_{SM}(x_s,y_s,x_t) = \sum_{k=1}^K \Phi(C^k_S,C^k_T),
\end{equation}
\textcolor{blue}{where $C^k_S$ and $C^k_T$ represent the centroid of class $K$ of the source and target domains respectively, and $\Phi(\cdot)$ represents the Euclidean distance between the two.}
This strategy is designed to mitigate the adverse effects of incorrect pseudo-labels. However, in situations with severe label shift, an excess of unreliable pseudo-labels can misalign centroids. We suggest that each sample’s contribution to the centroid calculation varies based on its reliability. For instance, in a binary classification problem, if samples $x_1$ and $x_2$ have probability outputs $[0.9, 0.1]$ and $[0.6, 0.4]$ respectively, $x_1$ is more reliable. Hence, we use confidence as a sample weight. For a sample $x$, the final probability output through a deep model parameterized by $\theta$ is represented as $p_{\theta} (y|x)$, with a weight of $w=\max p_{\theta}(y|x)$.

Inspired by contrastive learning ~\cite{hjelm2018learning}, the centroids with same class label but different domains should be closer, while the centroids with different class labels and domains should be futher away. We rewrite Eq.(\ref{eq:(2)}) as follows:
\begin{equation}\label{eq:(3)}
    L_{DSM}(x_s,y_s,x_t,\widehat y_t) = \frac{\sum_{k=1}^K \Phi(C^k_{wS},C^k_{wT})}{\sum_{i\neq k} \Phi(C^i_{wS},C^k_{wT})},
\end{equation}
where $C^k_{wS}$ and $C^k_{wT}$ represent the centroids weighted by $w$.

\subsubsection{Discriminative Feature Alignment}
% In the presence of two domains with large distribution disparities, our objective is to ensure their features are both domain-invariant and class-discriminative. Consequently, features sharing the same class label across different domains tend to align closely, while those with differing class labels across domains are distinctly separated. To advance this aim, we introduce discriminative feature alignment, which is based on contrastive learning. The method calculates the difference between each pair of features from the source and target domains. The source domain uses actual label values, while the target domain uses pseudo-labels produced by the classifier. When the class labels of the pair are identical, the features are drawn closer; if they differ, the features are pushed apart. This essentially constitutes cluster learning, which is used to establish more robust classification boundaries. Concurrently, to prevent unreliable samples from being overly attracted, we continue to use w as a sample’s weight. Formally,
In scenarios with two domains exhibiting significant distribution disparities, our goal is to ensure domain-invariant and class-discriminative features. Features with identical class labels across domains align closely, while those with different labels are distinctly separated. We propose discriminative feature alignment, a contrastive learning-based method, to facilitate this. It computes the difference between each feature pair from the source and target domains, using actual labels for the source and classifier-produced pseudo-labels for the target. Identical class labels draw features closer, while differing labels push them apart, effectively enabling cluster learning for robust classification boundaries. To avoid over-attracting unreliable samples, we persist in using w as a sample weight. Formally,

\begin{equation}\label{eq:(4)}
% \resizebox{.88\linewidth}{!}{$
%             \displaystyle
            L_{DFA}=\frac{{1}/{N_{same}}\sum_i \sum_j \sqrt{w_i^s w_j^t} \Phi(x_i^s,x_j^t) \rvert_{y_i^s=\widehat{y}_j^t}}{{1}/{N_{diff}}\sum_i \sum_j \sqrt{w_i^s w_j^t} \Phi(x_i^s,x_j^t) \rvert_{y_i^s \neq \widehat{y}_j^t}}
        % $}.
\end{equation}
% \begin{equation}\label{eq:(4)}
% \resizebox{.91\linewidth}{!}{$
%             \displaystyle
%             L_{DFA}=\frac{\sum_i \sum_j \sqrt{w_i^s w_i^t} d_{same}(x_i^s,x_j^t) \rvert_{y_i^s=\widehat{y}_j^t}}{\sum_i \sum_j \sqrt{w_i^s w_i^t} d_{diff}(x_i^s,x_j^t) \rvert_{y_i^s \neq \widehat{y}_j^t}}
%         $}.
% \end{equation}

The above strategies address covariate shift by aligning conditional distributions. However, when class imbalance is present, label shift becomes more pronounced, leading to a biased classifier and impacting the reliability of pseudo-labels. Given the unknown target domain, for generality and simplicity, we employ class-balanced sampling on the source domain. Specifically, when selecting samples for a mini-batch, each class has an equal probability of being selected.

\subsection{Label Shift Calibration (LSC)}

\subsubsection{Label Shift Metrics $M_{ls}$}
Label shift, quantifies the disparity in label distributions between source and target domains. It varies per class due to differing quantity distributions across domains. $M_{ls}$ is defined with respect to the probability distributions $P_S$ and $P_T$ of the source and target domains respectively.
% Label shift refers to the difference in label marginal distributions between the source and target domains. Due to the disparity in quantity distribution across two domains, the degree of label shift varies for each class. To measure this, we propose a label shift metric, denoted as $M_{ls}$. Assuming the probability distributions of the source and target domains are $P_S$ and $P_T$ respectively, then $M_{ls}$ can be expressed as:
\begin{equation}\label{eq:(5)}
    \textcolor{blue}{M_{ls}^{i}={P_T^i/P_S^i}}.
\end{equation}

$M_{ls}$, a $1\times C$ tensor, measures label shift, where $C$ is the number of classes and $M_{ls}^{i}$ represents the label shift degree for class $i$. If $\textcolor{blue}{M_{ls}^i}=1$, there's no label shift for class $i$. If $\textcolor{blue}{M_{ls}^i}>1$, class $i$ is more prevalent in the target domain, and if $0<\textcolor{blue}{M_{ls}^i}<1$, it's more prevalent in the source. Both cases indicate label shift, affecting pseudo-label reliability and potentially leading to error accumulation and performance degradation. We derive $P_S$ from source labels. The unlabeled target domain's $P_T$ is approximated using pseudo-labels, denoted as $\widehat P_T$.

% Given the source domain’s labels, we calculate $P_S$ using actual label counts. For unlabeled target domain, we apply the aforementioned conditional distribution alignment strategy for pre-training. This allows us to approximate $P_T$ withe pseudo-labels from the target, denoted as $\widehat P_T$.

% $M_{ls}$ is a $1\times C$ tensor, where $C$ denotes the number of classes, and $M_{ls}[i]$ signifies the degree of label shift for class $i$. (\romannumeral1) When $M_{ls}[i]=1$, it indicates no label shift for that class; (\romannumeral2) When $M_{ls}[i]>1$, it suggests class $i$ has a larger presence in the target domain than in the source domain; (\romannumeral3) When $0<M_{ls}[i]<1$, it implies class $i$ has a larger presence in the source domain than in the target domain. Scenarios (\romannumeral2) and (\romannumeral3) indicate a significant difference in label distribution between the source and target domains, i.e. label shift. The existence of label shift affects the reliability of pseudo-labels, and unreliable pseudo-labels can lead to error accumulation, thereby affecting the model’s classification performance.
\subsubsection{Label Shift Calibration (LSC)}

% Deep learning classification model can be decomposed  into a feature extractor $G$ and a classifier $F$. The goal of domain adaptation is to align the features extracted by $G$ from two domains. As the source data is labeled, the outcome of $F$ will fit the probability distribution of the source domain, i.e., $P_S$. In the case of long-tailed source data, $F$ tends to favor head classes with a quantity advantage, biasing against tail classes with fewer instances, leading to suboptimal learning of tail classes. When there is a significant difference between $P_S$ and $P_T$, the labeling result of $F$ for the target domain samples is unreliable. A simple but quite effective measure to solve this problem is to adopt class-balanced sampling for the source domain, compelling the neural network to learn from a uniform label marginal distribution, ensuring that$ F$ is class-unbiased. However, as $P_T$ is unknownand when the target domain follows a class-imbalanced long-tail distribution, the unbiased $F$ still cannot guarantee the reliability of the target sample labeling. Therefore, we propose LSC based on the degree of label shift $M_{ls}$, which calibrate the classification prediction for the target samples during the training process, making the labeling pseudo labels more consistent with the probability distribution of real target data to improve the reliability of target pseudo-labels. 

In deep learning classification models, we decompose them into a feature extractor $G$ and a classifier $F$. The goal of domain adaptation is to align the features extracted by $G$ from two domains. When dealing with long-tailed source data, $F$ tends to favor head classes due to their larger quantity, which can lead to suboptimal learning for tail classes with fewer instances. However, even if we adopt class-balanced sampling for the source domain, ensuring an unbiased $F$, the reliability of target sample labeling remains uncertain when the target domain follows a class-imbalanced long-tail distribution. To address this, we propose LSC based on the degree of label shift $M_{ls}$. LSC calibrates the classification predictions for target samples during training, making the pseudo labels more consistent with the real target data’s probability distribution, thus improving the reliability of target pseudo-labels.

For a target sample through a model with parameters $\theta$, we use $p_{\theta} (y|x_T)$ to represent its final probability output. The idea of LSC is to reweight $p_{\theta} (y|x_T)$ based on the degree of label shift $M_{ls}$, in order to re-estimate the target pseudo-labels. The class weighting matrix $W_m$ is designed as:
\begin{equation}\label{eq:(6)}
    W_m=\frac{1}{h_m+exp(-{\sqrt{M_{ls}}})}.
\end{equation}

Then we obtain target pseudo labels after calibration and its confidence weight:
\begin{equation}\label{eq:(7)}
    \widehat y_t^m=argmax \{p_{\theta}(y|x_T) \cdot W_m \}.
\end{equation}
\begin{equation}\label{eq:(8)}
    w^m= p_{\theta}(\widehat y_t^m|x_T) 
\end{equation}

A larger $M_{ls}[i]$ suggests that class $i$ is less frequent in the source but more so in the target domain, and vice versa for a smaller $M_{ls}[i]$. As per Eq.\ref{eq:(6)} and Eq.\ref{eq:(7)}, when a sample's feature is on the boundary of two classes and $M_{ls}[i] > M_{ls}[j]$, we prefer to label the sample as $i$, as shown in Figure 1. $W_m$ bounds the class weighting values, with $h_m$ set to 1.5, indicating that only unreliable samples at the classification boundary are calibrated to prevent over-calibration. The sample's confidence weight $w^m$ is determined by the classifier's output, mitigating the negative effects of incorrect classification calibration.

% When $M_{ls}[i]$ is larger, it indicates that class $i$ is less common in the source but more in the target domain. A smaller $M_{ls}[i]$ indicates the opposite. As per Eq.\ref{eq:(6)} and Eq.\ref{eq:(7)}, for $M_{ls}[i] > M_{ls}[j]$ and when a sample’s feature lies on the boundary of two classes, we favor labeling the sample as $i$, which is heuristically reasonable. It is illustrated in Figure 1. $W_m$ bounds the maximum and minimum values of class weighting , where $h_m$ is set to 1.5, signifying that only unreliable samples at the classification boundary are calibrated to avoid over-calibration. And the sample’s confidence weight $w^m$ is governed by the classifier’s output, thus reducing adverse effects of incorrect classification calibration.

\subsection{Overall Optimization and Analysis}

\subsubsection{Overall Optimization}
In summary, our training process comprises two stages. The first stage involves pre-training for three epochs, utilizing high-confidence target samples from the training results to estimate the target domain’s label distribution. The optimization objective of the first stage is:
\begin{align}\label{eq:(9)}
\resizebox{.9\linewidth}{!}{$
            % \displaystyle
    L_{total} = L_C(x_s,y_s)+
    \lambda L_{DSM}(x_s,y_s,x_t,\widehat y_t)+
   % \nonumber +\\
    \mu L_{DFA}(x_s,y_s,x_t,\widehat y_t)
    +\gamma L_{DC}(x_s,y_s)
    $}
\end{align}
In the second stage, we employ LSC to rectify target pseudo-labels $\widehat y_t^m$, and utilize $\widehat y_t^m$ for the training of CCA. Then our optimization objective is:
 \begin{align}\label{eq:(10)}
 \resizebox{.9\linewidth}{!}{$
            % \displaystyle
    L_{total}^m = L_C(x_s,y_s)+
    \lambda L_{DSM}(x_s,y_s,x_t,\widehat y_t^m)+
    \mu L_{DFA}(x_s,y_s,x_t,\widehat y_t^m)
    +\gamma L_{DC}(x_s,y_s)
    $}
\end{align}
where $\lambda$ and $\mu$ and $\gamma$ are hyperparameters no less than zero. 
% The optimization procedure is shown in Algorithm 1.

% \begin{algorithm}[tb]
%     \caption{CCA-LSC ptimization}
%     \label{alg:algorithm}
%     \textbf{Input}:Labeled source examples $(x^s,y^s)$, unlabeled target data $(x^t)$, parameters $\lambda$ and $\mu$ and $\gamma$
%     \begin{algorithmic}[1] %[1] enables line numbers
%         \FOR{epoch in Pre\_EPOCHS}
%         \STATE Update network model parameters $\theta$ by minimizing losses in Eq.\ref{eq:(9)}
%         \ENDFOR 
%         \STATE Estimated $P_S$ via the source's real label $y_s$. Estimate $\widehat P_T$ via pseudo labels $\widehat y_t$ with a target sample confidence level $w>0.5$ and calculate the class weighting matrix $W_m$ via Eq.\ref{eq:(5)} and Eq.\ref{eq:(6)}.
%         \FOR{epoch in range (Pre\_EPOCHS, MAX\_EPOCHS)}
%         \FOR{each iteration}
%         \STATE Update $\widehat y_t^m$ via Eq.\ref{eq:(7)}
%         \STATE Recalculate the conditional alignment loss with $\widehat y_t^m$ via Eq.\ref{eq:(3)} and Eq.\ref{eq:(4)}
%         \STATE Update network model parameters $\theta$ by minimizing losses in Eq.\ref{eq:(10)}
%         \ENDFOR
%         \ENDFOR 
%     \end{algorithmic}
% \end{algorithm}

\subsubsection{Analysis}
Next, we demonstrate how our approach reduces the expected error on the target samples from domain adaptation theory.

\begin{theorem}[~\cite{ben2010theory}]\label{eq:(11)}
    Denote $h \in \mathcal{H}$ as the hypothesis. Given two domains $\mathcal{S}$ and $\mathcal{T}$, the target error $\varepsilon_{\mathcal{T}}$ is bounded by three terms: (\romannumeral1) $\varepsilon_{\mathcal{S}}$: source error, (\romannumeral2) $d_{\mathcal{H}\Delta \mathcal{H}}(\mathcal{S},\mathcal{T})$: the discrepancy distance between two distributions S and T, (\romannumeral3) $C_0$:shared expected loss. We have:
    \begin{equation}
        \forall h \in \mathcal{H}, \varepsilon_{\mathcal{T}}(h) \leq \varepsilon_{\mathcal{S}}(h) + \frac{1}{2}d_{\mathcal{H}\Delta \mathcal{H}}(\mathcal{S},\mathcal{T}) + C_0.
    \end{equation}
\end{theorem}

It is defined as $C_0 = \min_{h \in \mathcal{H}} \varepsilon_{\mathcal{S}}(h, f_{\mathcal{S}}) + \varepsilon_{\mathcal{T}}(h, f_{\mathcal{T}})$ where $f_{\mathcal{S}}$ and $f_{\mathcal{T}}$ are labeling functions for source and target domain respectively. 
Previous methods often assume that $C_0$ is negligible. However, when $C_0$ is large, ignoring $C_0$ can prevent the learning of an effective target classifier.

\begin{theorem}[~\cite{xie2018learning}]\label{eq:(12)}
    According to the triangle inequality for classification error~\cite{ben2010theory,crammer2008learning}, an upper bound for $C_0$ is:
    \begin{equation}
         % \resizebox{.91\linewidth}{!}{$
         %    \displaystyle
                \begin{split}
                C_0 = \min_{h \in \mathcal{H}} & \varepsilon_{\mathcal{S}}(h, f_{\mathcal{S}}) + \varepsilon_{\mathcal{T}}(h, f_{\mathcal{T}}) 
                \\
                \leq \min_{h \in \mathcal{H}} & \varepsilon_{\mathcal{S}}(h, f_{\mathcal{S}}) + \varepsilon_{\mathcal{T}}(h, f_{\mathcal{S}})
                +\varepsilon_{\mathcal{T}}(f_{\mathcal{S}},f_{\mathcal{T}})
                \\
                \leq \min_{h \in \mathcal{H}} & \varepsilon_{\mathcal{S}}(h, f_{\mathcal{S}}) + \varepsilon_{\mathcal{T}}(h, f_{\mathcal{S}})
                +\varepsilon_{\mathcal{T}}(f_{\mathcal{S}},f_{\widehat\mathcal{T}}) 
                 + \varepsilon_{\mathcal{T}}(f_{\mathcal{T}},f_{\widehat\mathcal{T}})
               \end{split}
            % $}.
    \end{equation}
\end{theorem}
In the given formula, the first two terms quantify the discrepancy between the hypothesis $h$ and the source labeling function $f_\mathcal{S}$. Given the availability of source labels, these terms are typically minimal, facilitating the learning of a hypothesis space $h$ that closely approximates $f_\mathcal{S}$. The third term measures the inconsistency between the source and pseudo-target labeling functions on target samples, while the final term indicates the divergence between the pseudo-target labeling function and the true target label, serving as a reliability measure for the pseudo-labels. Our method seeks to minimize the last two terms to optimize the upper bound of $C_0$.
The moving average centroid alignment strategy, discussed in ~\cite{xie2018learning}, optimizes the third term by aligning the centroids of target and source features in class $C_0$, ensuring prediction consistency. Our approach employs both sample-weighted moving average centroid alignment and discriminative feature alignment to foster feature alignment across different domains but within the same class, thereby minimizing the third term.

However, ~\cite{xie2018learning} presumes the fourth term will minimize over time and disregards it. This assumption falls short in the presence of data imbalance and label shift, where optimizing the third term could induce class bias in the pseudo-target labeling function, amplifying the fourth term. Our proposed LSC rectifies this by adjusting the classification prediction of the pseudo-target labeling function based on the label shift index $M_{ls}$, reducing the false pseudo-rate, and aligning the prediction with the true target data’s label distribution, thereby also minimizing the fourth term. Our experiments demonstrate that LSC consistently curtails the false pseudo-rate on target samples (refer to section 4.4).

In essence, the efficacy of domain adaptation methods hinges on managing each term that could escalate the target classification error, thus broadening the applicability of domain adaptation methods.

\section{Experiments}

\subsection{Set up}
\subsubsection{Datasets}
We utilized three datasets . First, we employed {\bf Office-Home (RS-UT)}, an imbalanced version of Office-Home created by ~\cite{tan2020class}, where the source and target domains follow two reverse Paredo distributions. This benchmark includes three domains: Clipart (Cl), Product images (Pr), and Real-world images (Rw). The Art images (Ar) domain in Office-Home, being too small for sampling an imbalanced subset, is not considered here. Second, we used a subset of {\bf DomainNet} created by ~\cite{tan2020class}, which includes 40 classes from four domains (Real (R), Clipart (C), Painting (P), Sketch (S)). As a noticeable label shift already exists, we made no additional modifications. The label distributions can be seen in Figure \ref{dataset}. \textcolor{blue}{{\bf Office-31} ~\cite{office31} contains 4,110 images of 31 categories. The domains are Amazon (A), Webcam (W), and DSLR (D).}

\subsubsection{Baselines}
We benchmarked our method against eight state-of-the-art techniques that tackle both covariate shift and label shift. (\romannumeral1) {\bf COAL} ~\cite{tan2020class} aligns feature and label distributions using prototype-based conditional alignment and self-training on confident pseudo-labels. (\romannumeral2) {\bf MDD+Implicit Alignment (I.A)} ~\cite{jiang2020implicit} removes explicit model parameter optimization from pseudo-labels via sampled implicit alignment. (\romannumeral3) {\bf InstaPBM} ~\cite{li2020rethinking} employs instance-based prediction behavior matching. (\romannumeral4) {\bf F-DANN} ~\cite{wu2019domain} introduces a DANN based on asymmetric relaxed distribution matching. (\romannumeral5) {\bf SENTRY} ~\cite{prabhu2021sentry} minimizes the entropy of reliable instances and maximizes that of unreliable ones. \textcolor{blue}{(\romannumeral6){\bf TIToK}~\cite{TIToK} and (\romannumeral7){\bf BIWAA-I}~\cite{BIWAA-I} and (\romannumeral8){\bf RHWD}~\cite{RHWD} also solve both label and feature shifting problems.} All methods, except F-DANN, use target pseudo-labels. We also compared with conventional UDA methods like BBSE~\cite{lipton2018detecting}, which only addresses label shift, and MCD~\cite{saito2018maximum}, DAN~\cite{long2015learning}, DANN~\cite{ganin2016domain}, JAN~\cite{long2017deep}, BSP~\cite{chen2019transferability}, which solely focus on covariate shift.

\subsubsection{ Implementation details}
All experiments are conducted using the Pytorch framework with resnet50. The model's hyper-parameters are $\lambda =3$, $\mu =0.6$, and $\gamma =1$. The bottleneck layer dimension is 256, and the batch size is 50. We use the SGD optimizer with a momentum of 0.9. The initial learning rate for the classifier is 0.005 for OfficeHome and 0.01 for DomainNet, adjusted as ~\cite{ganin2016domain}. The model trains for 20 epochs, with the first 3 forming the initial stage. After this, the model evaluates target samples and uses pseudo-labels with a confidence level of $w > 0.5$ to estimate the target domain's label distribution. The model then enters the second stage. The random seed is set to 100 for reproducibility. For imbalanced data, we use per-class mean accuracy, as suggested by ~\cite{tan2020class}, for a fair performance assessment.

% We conduct all experiments using the Pytorch framework and resnet50 CNN architecture as the backbone. We specify the following hyper-parameters for our model: $\lambda =3$, $\mu =0.6$ and $\gamma =1$. We set the bottleneck layer dimension to 256, batch size to 50, and use Stochastic Gradient Descent optimizer (SGD) with momentum 0.9. On OfficeHome and DomainNet, the classifier's initial learning rate is set to 0.005 and 0.01 respectively, and its value is changed according to ~\cite{ganin2016domain}. Our model trains 20 epochs, with the initial 3 epochs constituting the first stage of training. Upon completion of this stage, the model evaluates the target samples and utilizes pseudo-labels with a confidence level of $w > 0.5$ to estimate the label distribution of the target domain. Subsequently, the model enters the second stage of training. Throughout this process, the random seed is set to 100 to ensure reproducibility. In the context of imbalanced data, traditional accuracy metrics may not accurately reflect performance. Therefore, we choose to use the per-class mean accuracy, as proposed by ~\cite{tan2020class}, as our evaluation metric. This approach ensures a more fair assessment of our model’s performance across all classes.

\begin{table}[ht]
\centering
% \small
% \begin{tabular}{@{}lccccccccccccc@{}}
% \begin{footnotesize}
\begin{adjustbox}{width=1.\textwidth}
\begin{tabular}{@{}lccccccccccccc@{}}
% \begin{tabular}{@{}l@{}c @{}c @{}c @{}c @{}c @{}c @{}c @{}c @{}c @{}c @{}c @{}c @{}c @{} }
\toprule
Methods & R→C & R→P & R→S & C→R & C→P & C→S & P→R & P→C & P→S & S→R & S→C &S→P & AVG \\ \midrule
source & 65.75 & 68.84 & 59.15 & 77.71 & 60.60 & 57.87 & 84.45 & 62.35 & 65.07 & 77.10 & 63.00 & 59.72 & 66.80 \\ \midrule
% BBSE   &55.38 & 63.62 & 47.44 & 64.58 & 42.18 & 42.36 & 81.55 & 49.04 & 54.10 & 68.54 & 48.19 & 46.07 & 55.25\\
MCD    &61.97 & 69.33 & 56.26 & 79.78 & 56.61 & 53.66 & 83.38 & 58.31 & 60.98 & 81.74 & 56.27 & 66.78 & 65.42 \\
% DAN    &64.36 & 70.65 & 58.44 & 79.44 & 56.78 & 60.05 & 84.56 & 61.62 & 62.21 & 79.69 & 65.01 & 62.04 & 67.07 \\
DANN   &63.37 & 73.56 & 72.63 & 86.47 & 65.73 & 70.58 & 86.94 & 73.19 & 70.15 & 85.73 & 75.16 & 70.04 & 74.46 \\
F-DANN &66.15 & 71.80 & 61.53 & 81.85 & 60.06 & 61.22 & 84.46 & 66.81 & 62.84 & 81.38 & 69.62 & 66.50 & 69.52\\
JAN    &65.57 & 73.58 & 67.61 & 85.02 & 64.96 & 67.17 & 87.06 & 67.92 & 66.10 & 84.54 & 72.77 & 67.51 & 72.48\\
BSP    &67.29 & 73.47 & 69.31 & 86.50 & 67.52 & 70.90 & 86.83 & 70.33 & 68.75 & 84.34 & 72.40 & 71.47 & 74.09\\
COAL   &73.85 & 75.37 & 70.50 & 89.63 & 69.98 & 71.29 & 89.81 & 68.01 & 70.49 & 87.97 & 73.21 & 70.53 & 75.89 \\
MDD+I.A &78.54 & 75.09 & 69.43 & 88.50 & 70.59 & 70.44 & 88.37 & 75.71 & 71.65 & 89.35 & 77.97 & 72.41 & 77.33 \\
InstaPBM & 80.10 & 75.87 & 70.84 & 89.67 & 70.21 & 72.76 & 89.60 & 74.41 & 72.19 & 87.00 & 79.66 & 71.75 & 77.84\\
SENTRY & {83.89} & 76.72 & 74.43 & { 90.61} & 76.02 &{79.47} & 90.27 & {82.91} & 75.60 & 90.41 & 82.40 & 73.98 & 81.39 \\ 
\textcolor{blue}{BIWAA-I} &
\textcolor{blue}{79.93} &
\textcolor{blue}{75.24} &
\textcolor{blue}{75.35} &
\textcolor{blue}{87.93} &
\textcolor{blue}{72.07} &
\textcolor{blue}{75.71} &
\textcolor{blue}{88.87} &
\textcolor{blue}{77.81} &
\textcolor{blue}{76.66} &
\textcolor{blue}{88.78} &
\textcolor{blue}{80.49} &
\textcolor{blue}{74.49} &
\textcolor{blue}{79.44} \\

\textcolor{blue}{RHWD ~\cite{RHWD}} &
\textcolor{red}{84.80} &
\textcolor{blue}{76.90} &
\textcolor{blue}{75.20} &
\textcolor{red}{91.80} &
\textcolor{blue}{75.60} &
\textcolor{red}{81.20} &
\textcolor{red}{91.90} &
\textcolor{red}{84.60} &
\textcolor{blue}{76.10} &
\textcolor{red}{91.30} &
\textcolor{red}{83.20} &
\textcolor{blue}{74.60} &
\textcolor{blue}{82.00}  \\

\midrule
\small Ours  & 83.74 & \textcolor{red} {77.10} & \textcolor{red} {79.00} & 90.21 & \textcolor{red} {76.54} & 78.55 & 89.62 & 81.86 & \textcolor{red} {79.57} & {90.49} & {83.06}
& \textcolor{red} {77.48} & \textcolor{red} {82.27}\\
\bottomrule
\end{tabular}
% \end{footnotesize}
\end{adjustbox}
\caption{Per-class average accuracies on DomainNet}
\label{domainnet}
% \end{table}
% \begin{table}[ht]
\centering
\begin{adjustbox}{width=0.75\textwidth}
\begin{tabular}{@{}lccccccc@{}}
% \small
% \begin{tabular}{@{}l@{}c @{}c @{}c @{}c @{}c @{}c @{  }c @{} }
\toprule
Methods & Rw$\to$Pr & Rw$\to$Cl & Pr$\to$Rw & Pr$\to$Cl & Cl$\to$Rw & Cl$\to$Pr &AVG  \\ \midrule
source & 70.74 & 44.24 & 67.33 & 38.68 & 53.51 & 51.85 & 54.39 \\ \midrule
BBSE   &61.10 & 33.27 & 62.66 & 31.15 & 39.70 & 38.08 & 44.33\\
MCD    &66.03 & 33.17 & 62.95 & 29.99 & 44.47 & 39.01 & 45.94\\
DAN    &69.35 & 40.84 & 66.93 & 34.66 & 53.55 & 52.09 & 52.90\\
DANN   &71.62 & 46.51 & 68.40 & 38.07 & 58.83 & 58.05 & 56.91\\
F-DANN &68.56 & 40.57 & 67.32 & 37.33 & 55.84 & 53.67 & 53.88\\
JAN    &67.20 & 43.60 & 68.87 & 39.21 & 57.98 & 48.57 & 54.24\\
COAL   &73.65 & 42.58 & 73.26 & 40.61 & 59.22 & 57.33 & 58.40\\
MDD+I.A &76.08 & 50.04 & 74.21 &  45.38 &  61.15 & 63.15 & 61.67\\
InstaPBM & 75.56 & 42.93 & 70.30 & 39.32 & 61.87 & 63.40 & 58.90\\
SENTRY & 76.12 & 56.80 & 73.60 & \textcolor{red} {54.75} & \textcolor{red} {65.94} & 64.29 & 65.25 \\ 
\textcolor{blue}{TIToK} &
\textcolor{blue}{77.09} &
\textcolor{blue}{52.84} &
\textcolor{blue}{72.15} &
\textcolor{blue}{44.32} &
\textcolor{blue}{60.06} &
\textcolor{blue}{59.95} &
\textcolor{blue}{61.07} \\

\midrule
Ours  & \textcolor{red} {79.18} & \textcolor{red} {60.53} & \textcolor{red} {78.26} & 50.13 & 65.79 & \textcolor{red}{ 68.99} & \textcolor{red} {67.15} \\
\bottomrule
\end{tabular}
\end{adjustbox}
\caption{Per-class average accuracies on OfficeHome (RS-UT)}
\label{officehome}
\end{table}

\begin{table}[ht]
\centering
\begin{adjustbox}{width=0.66\textwidth}
\begin{tabular}{@{}lccccccc@{}}
% \small
% \begin{tabular}{@{}l@{}c @{}c @{}c @{}c @{}c @{}c @{  }c @{} }
\toprule
\textcolor{blue}{Methods} & \textcolor{blue}{A}$\to$\textcolor{blue}{W} & \textcolor{blue}{D}$\to$\textcolor{blue}{W} & \textcolor{blue}{W}$\to$\textcolor{blue}{D} & \textcolor{blue}{A}$\to$\textcolor{blue}{D} & \textcolor{blue}{D}$\to$\textcolor{blue}{A} & \textcolor{blue}{W}$\to$\textcolor{blue}{A} &\textcolor{blue}{AVG}  \\ \midrule
% source & 70.74 & 44.24 & 67.33 & 38.68 & 53.51 & 51.85 & 54.39 \\ \midrule
\textcolor{blue}{DAN} & 68.5& 96.0& 99.0 &67.0 &54.0& 53.1& 72.9\\
\textcolor{blue}{DANN} & 82.0& 96.9 &99.1& 79.7 &68.2 &67.4& 82.2\\
\textcolor{blue}{MCD} &88.6 &98.5 &100.& 92.2& 69.5& 69.7 &86.5\\
\textcolor{blue}{MDD} & 94.5 & 98.4& 100. &93.5 &74.6 &72.2 &88.9 \\ 
\textcolor{blue}{BIWAA-I}& 95.6& 99.0 &100. &\textcolor{red}{95.4}& 75.9 &77.3 &90.5\\

\midrule
\textcolor{blue}{Ours}  & \textcolor{red} {96.0} & \textcolor{red} {99.1} & \textcolor{red} {100.} & {94.6} & \textcolor{red}{77.1} & \textcolor{red}{ 77.3} & \textcolor{red} {90.7} \\
\bottomrule
\end{tabular}
\end{adjustbox}
\caption{\textcolor{blue}{Accuracy results on Office-31 dataset.}}
\label{office31}
\end{table}

\subsection{Results}

\begin{figure*}[h]
\centering
\subfigure[]{
\includegraphics[width=0.22\textwidth]{figure/different label shift_1.jpg}
\label{different label shift}
}
\subfigure[]{
\includegraphics[width=0.22\textwidth]{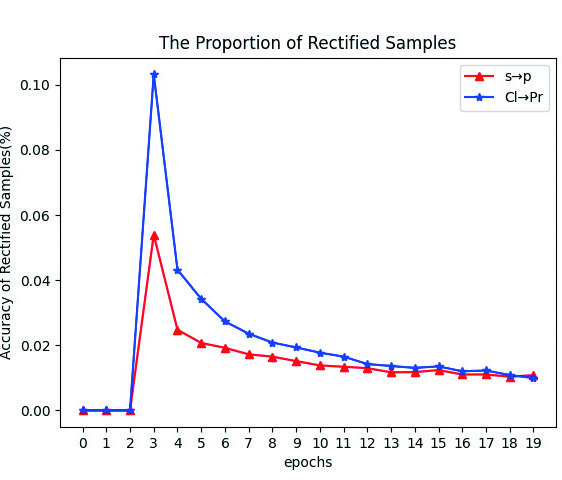}
\label{proportion}
}
\subfigure[]{
\includegraphics[width=0.22\textwidth]{figure/Cl2Pr _1.jpg}
\label{Cl2Pr}
}
\subfigure[]{
\includegraphics[width=0.22\textwidth]{figure/s2p _1.jpg}
\label{s2p}
}
\caption{Analysis of label shift calibration. (a) Accuracy under different degrees of imbalance on Cl$\to$Pr. (b) The proportion of target samples with calibrated pseudo labels($\widehat y \neq \widehat y^m$) via CCA-LSC. (c) and (d) :The accuracy of the target pseudo labels $\widehat y$ (obtained by the classifier) and $\widehat y^m$ (calibrated based on label shift metric $M_{ls}$) in all calibrated target samples ($\widehat y \neq \widehat y^m$) on Cl$\to$Pr and S$\to$P respectively.  }
\label{fig:mainfig}
\end{figure*}

% \subsubsection{DomainNet and OfficeHome}
{\bf DomainNet and OfficeHome.}
The experimental results on DomainNet and OfficeHome are presented in Tables \ref{domainnet} and \ref{officehome}, respectively. Our method outperforms the second best method SENTRY, by improving the average accuracy by 1.90\% on OfficeHome (RS-UT) and by 0.88\% on DomainNet. Table \ref{domainnet} reveals that our method significantly surpasses SENTRY in scenarios with higher label shifts, such as R$\to$S, P$\to$S, and S$\to$P, registering increases of 4.57\%, 3.97\%, and 3.50\%, respectively. Table \ref{officehome} shows a better promotion since there are severe label shift. These results highlight our method's efficacy in simultaneously tackling label shift and covariate shift. \\
\textcolor{blue}{{\bf Office-31.}
The experimental results are shown in Tables \ref{office31}. There are few label shifts but feature shifts in this dataset. It can be seen that our method also has good performance for solving the problem of feature shifting.}\\
{\bf Different Degrees of Label Shift.}
% \subsubsection{Different Degrees of Label Shift}
We measure imbalance using the imbalance factor IF~\cite{cui2019class}, defined as the ratio of maximum to minimum class sizes. A larger IF indicates more imbalance. We created four splits on Cl$\to$Pr with IF$\in\{1,5,10,20\}$. For IF=1, we used the original Cl and Pr data from OfficeHome. For other splits, we maintained the maximum class size and adjusted the Pareto distribution parameters based on OfficeHome (RS-UT). All methods used class-balanced sampling in the source domain for fairness. As shown in Figure \ref{different label shift}, accuracy decreases for all methods with increasing imbalance due to label shift, but our method consistently outperforms the others.

% We use the imbalance factor IF~\cite{cui2019class}, defined as the ratio of maximum to minimum class cardinalities, to measure imbalance. A higher IF indicates greater imbalance. Four splits on Cl$\to$Pr were created with IF$\in\{1,5,10,20\}$. For IF=1, we employed the original Cl and Pr data from OfficeHome. For other splits, we kept the maximum class cardinality constant and adjusted the Pareto distribution parameters based on OfficeHome (RS-UT). To ensure fairness, all methods used class-balanced sampling in the source domain. Figure \ref{different label shift} presents the experimental results. With increasing imbalance, accuracy decreases for all methods due to label shift, but our method consistently outperforms the others.

\subsection{Ablation Study}
To mitigate the influence of source data imbalance, we evaluated each domain adaptation component using class-balanced sampling on the source domain. Table \ref{ablation} presents the results. Model performance is bad with only source cross-entropy loss. Performance improves with the addition of adversarial learning and sample-weighted moving average centroid alignment loss ($L_{DC}$+$L_{DSM}$). Significant improvement is observed with the inclusion of discriminative feature alignment loss ($L_{DFA}$), which ensures both domain invariance and class discriminability of the learned representation. Label shift calibration on target samples further enhances performance by reducing the target false pseudo rate during training, ensuring correct execution of the two pseudo label-based strategies.

\begin{table}[h]
\centering

\begin{minipage}{.6\textwidth}
\flushleft
\begin{tabular}{@{}lccc@{}}
\toprule
Methods & Cl$\to$Pr & S$\to$P & P$\to$S\\ \midrule
source & 51.85 & 63.00 & 65.07\\ \midrule
$L_{C}$+$L_{DC}$ & 62.23 & 71.22 & 73.32\\
$L_{C}$+$L_{DC}$+$L_{DSM}$ & 63.37 & 72.80 & 75.08\\
$L_{C}$+$L_{DC}$+$L_{DSM}$+$L_{DFA}$ & 66.57 & 76.74 & 77.35\\
$L_{C}$+$L_{DC}$+$L_{DSM}^m$+$L_{DFA}^m$ & {\textcolor{red} {68.99}} & {\textcolor{red} {77.48}} & {\textcolor{red} {79.57}}\\
\bottomrule
\end{tabular}
\caption{\textcolor{blue}{Ablation Study: effectiveness of adap-\\tation components. We adopt class-balanced\\ sampling on the source domain to counteract\\ the adverse effect caused by imbalance to ex-\\amine the effectiveness of each component.}}
\label{ablation}
\end{minipage}%
\begin{minipage}{.35\textwidth}
\centering
% \small
\begin{tabular}{@{}lccc@{}}
\toprule
$h_m$         & 1  &  1.5  &  2 \\  \midrule
Cl$\to$Pr   & 68.32 & {\textcolor{red} {68.99}} & 68.44\\ 
\bottomrule
\end{tabular}
\caption{The influence of $h_m$}
\label{tab:$h_m$}

\begin{tabular}{@{}c|ccc@{}}
\toprule
\diagbox[]{$\lambda$}{$\mu$} & 0.4 & 0.6 & 0.8   \\  \midrule
1         & 67.18  & 67.73   & 68.06  \\  
3         & 68.22  &{\textcolor{red} {68.99}}    & 68.34  \\   
5         & 68.56  & 68.11  & 67.16 \\
\bottomrule
\end{tabular}
\caption{Hyper-parameter sensitivity on Cl$\to$Pr}
\label{tab:Hyper-parameter}
\end{minipage}
\end{table}

\subsection{Analysis of Label Shift Calibration}

The label shift calibration strategy calibrates only some target pseudo labels at the classification boundary, leading to two scenarios: consistency ($\widehat y=\widehat y^m$) and inconsistency ($\widehat y \neq \widehat y^m$) between the classifier’s output pseudo labels and the calibrated ones. Figure \ref{proportion} illustrates the proportion of samples with calibrated pseudo labels ($\widehat y \neq \widehat y^m$) during training, which decreases over time, indicating an increasing number of samples moving away from the classification boundary. Figures \ref{Cl2Pr} and \ref{s2p} show the right proportion of $\widehat y$ and $\widehat y^m$ in these calibrated samples. During the initial 3 epochs of pre-training, label shift calibration is not applied. Throughout the training, the accuracy of $\widehat y^m$ consistently surpasses that of $\widehat y$, demonstrating the strategy’s effectiveness in reducing the false pseudo rate of the classifier’s target output, supporting the analysis in Section 3.4. In fact, higher accuracy of $\widehat y^m$ over $\widehat y$, was observed in all 18 transfer tasks on OfficeHome and DomainNet during training.

A question naturally arises: given the label shift between source and target domains, could we diminish this shift by implementing pseudo-label balanced sampling on the target domain and class-balance sampling on the source domain? Initially, the balanced sampling strategy curbs imbalance by regulating the utilization of input data, inevitably leading to an over-sampling of certain classes. This is more likely to negatively impact the quality of the learned representation for unlabeled target domain data. Furthermore, our application of pseudo-label balanced sampling on the OfficeHome dataset resulted in a reduction of per-class accuracy by about 1\%. Consequently, we have decided not to use  pseudo-label balanced sampling strategy on target data in our method.

\subsection{Hyper-parameter Discussion}
{\bf The Influence of the Parameter $h_m$.}
The $h_m$ dictates the proportion of calibrated samples. A smaller $h_m$ value leads to a larger proportion of $\widehat y \neq \widehat y^m$. Although our calibration strategy is effective, more calibrations aren’t always better. Over-calibration can lead to over-representation of the dominant class in target samples, while under-calibration can lessen its effectiveness. Table \ref{tab:$h_m$} illustrates the impact of the $h_m$. To counteract the effects of incorrect calibrations, we derive the confidence of all target pseudo labels from the classifier’s probability output. For instance, if a sample’s probability output is [0.6, 0.4] and the calibrated output is [0.45, 0.55], its confidence is 0.4 and its weight $w=0.4$. This can effectively reduce the adverse impact of incorrect calibrations.\\
% \begin{table}[h]
% \centering
% \begin{minipage}{.4\textwidth}
% \centering
% % \small
% \begin{tabular}{@{}lccc@{}}
% \toprule
% $h_m$         & 1  &  1.5  &  2 \\  \midrule
% Cl$\to$Pr   & 68.32 & {\bf 68.99} & 68.44\\ 
% % $h_m$  &   Cl$\to$Pr \\ \midrule
% % 1      &    68.32  \\
% % 1.5    &  {\bf 68.99}\\
% % 2      &    68.44\\
% \bottomrule
% \end{tabular}
% \caption{The influence of the\\ parameter $h_m$}
% \label{tab:$h_m$}
% \end{minipage}%
% \begin{minipage}{.5\textwidth}
% \centering
% % \small
% \begin{tabular}{@{}c|ccc@{}}
% \toprule
% \diagbox[]{$\lambda$}{$\mu$} & 0.4 & 0.6 & 0.8   \\  \midrule
% 1         & 67.18  & 67.73   & 68.06  \\  
% 3         & 68.22  &{\bf 68.99}    & 67.34  \\   
% 5         & 67.56  &67.93  & 66.86 \\
% \bottomrule
% \end{tabular}
% \caption{Hyper-parameter sensitivity\\ on Cl$\to$Pr}
% \label{tab:Hyper-parameter}
% \end{minipage}
% \end{table}
{\bf{Hyper-parameter Analysis.}}
We fixed $\gamma$ to 1 and discussed the impact of $\lambda$ and $\mu$. The experimental results are shown in the Table \ref{tab:Hyper-parameter}. It can be seen that our experimental results are not sensitive to each hyperparameter.

\subsection{Analysis of Two Stage Learning}
\begin{figure*}[ht]
\centering
\subfigure[]{
\includegraphics[width=0.45\textwidth]{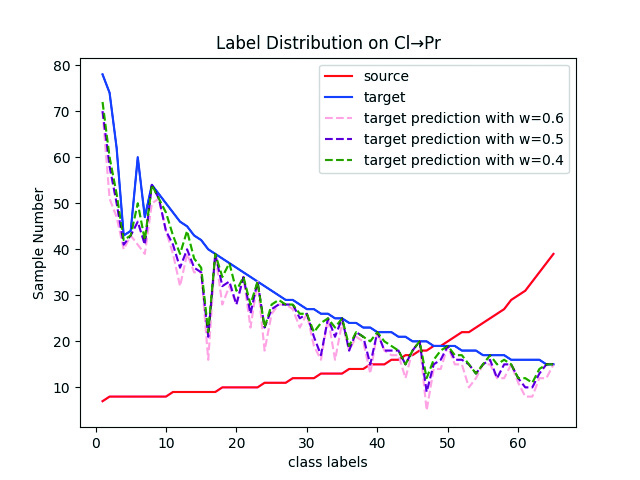}
\label{label_distribution_Cl_Pr}
}
\subfigure[]{
\includegraphics[width=0.45\textwidth]{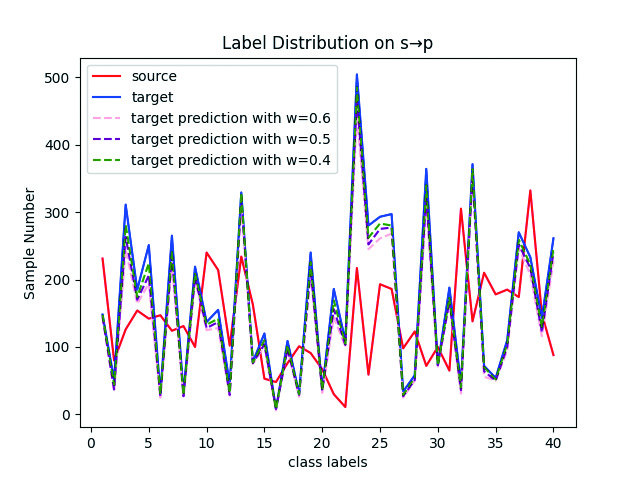}
\label{label_distribution_s_p}
}
\caption{The impact of selecting target samples with different confidence levels on the estimation of target label distribution}

\label{fig:mainfig3}
\end{figure*}
Our LSC strategy relies on the distribution estimation of the target domain in the first stage. When this estimation is highly unreliable, the LSC strategy may fail. Therefore, we discuss the impact of the pre-training of CCA in the first stage on the LSC strategy in the second stage. Figure~\ref{fig:mainfig3} shows the results of estimating the target domain distribution by selecting pseudo-labels of target samples with different confidence levels. It can be seen that when the confidence level $w > 0.4$, $w > 0.5$, $w > 0.6$, our estimated distribution of the target domain is generally close to its true distribution, indicating that $\widehat P_T$ is reliable. In fact, our estimation of the target domain distribution does not need to be very accurate, as long as it can generally reflect the target label distribution. In our experiments, we use pseudo-labels of target samples with a confidence level of $w > 0.5$ to estimate the target domain distribution.

 % Consider a simple example where the source domain A has two categories with quantities [100,10], and the target domain B has [10,100]. According to the formula in LSC, for the distribution estimation [m,n] of B, our LSC strategy is effective when m/100 < n/10. This condition is easily met. 

\section{Conclusion}
We introduce CCA-LSC to tackle label shift and covariate shift in imbalanced domain adaptation. Our approach employs domain adversarial learning, sample-weighted moving average centroid alignment, and discriminative feature alignment for contrastive conditional alignment, facilitating the learning of feature representations that are both domain-invariant and class-discriminative. To counter label shift, we introduce the label shift measure $M_{ls}$, using it to calibrate the classification prediction of target samples. Experimental evidence demonstrates that CCA-LSC delivers state-of-the-art results on benchmark datasets. \\
% \section*{Acknowledgements}
{\bf Acknowledgements.}
The paper is supported by the National Natural Foundation Science of China (62101061).

% \begin{table}
%     \centering
%     \begin{tabular}{lrr}
%         \toprule
%         Scenario  & $\delta$ (s) & Runtime (ms) \\
%         \midrule
%         Paris     & 0.1          & 13.65        \\
%                   & 0.2          & 0.01         \\
%         New York  & 0.1          & 92.50        \\
%         Singapore & 0.1          & 33.33        \\
%                   & 0.2          & 23.01        \\
%         \bottomrule
%     \end{tabular}
%     \caption{Booktabs table}
%     \label{tab:booktabs}
% \end{table}

%% The file named.bst is a bibliography style file for BibTeX 0.99c
% \small
% \bibliographystyle{named}
% \bibliography{article}
\small
\bibliographystyle{splncs04}

\end{document}